\documentclass[sigconf]{acmart}
\newcommand{\system}{\textsf{FactCheck Editor}}
\usepackage[normalem]{ulem}

\AtBeginDocument{%
  \providecommand\BibTeX{{%
    \normalfont B\kern-0.5em{\scshape i\kern-0.25em b}\kern-0.8em\TeX}}}

\copyrightyear{2024} 
\acmYear{2024} 
\setcopyright{acmcopyright} 

\acmConference[SIGIR '24]{Proceedings of the 47th International ACM SIGIR Conference on Research and Development in Information Retrieval}{July 14--18, 2024}{Washington, D.C., USA}

\acmISBN{978-1-4503-XXXX-X/18/06}

\usepackage{booktabs} 
\usepackage{color}

\begin{document}

\fancyhead{}
\title{FactCheck Editor: Multilingual Text Editor with End-to-End fact-checking} 

\author{Vinay Setty}
\affiliation{%
  \institution{University of Stavanger and Factiverse AI}
  \country{Stavanger}
  \country{Norway}
}
\email{vsetty@acm.org}

\begin{abstract}
We introduce \system{}, an advanced text editor designed to automate fact-checking and correct factual inaccuracies. Given the widespread issue of misinformation, often a result of unintentional mistakes by content creators, our tool aims to address this challenge. It supports over 90 languages and utilizes transformer models to assist humans in the labor-intensive process of fact verification. This demonstration showcases a complete workflow that detects text claims in need of verification, generates relevant search engine queries, and retrieves appropriate documents from the web. It employs Natural Language Inference (NLI) to predict the veracity of claims and uses LLMs to summarize the evidence and suggest textual revisions to correct any errors in the text. Additionally, the effectiveness of models used in claim detection and veracity assessment is evaluated across multiple languages.
\end{abstract}

\ccsdesc[500]{Information systems~Retrieval tasks and goals}

\keywords{Multilingual; Fact-checking}

\maketitle

\section{Introduction}

In the era of digital information, the proliferation of misinformation has emerged as a formidable challenge, impacting societies, politics, and public opinions. This is often a result of unintentional mistakes by content creators, has necessitated the development of tools that can effectively identify and correct factual errors~\cite{Guo:2022:JACL,Zhou:2018:a}.

Most newsrooms rely on content management systems for news production, offering basic formatting and composition tools. After journalists write an article, it is typically proofread and fact-checked manually, using web searches and searching internal archives. Current automation extends only to grammar checkers like Grammarly and advanced tools like Writer.com, which automate writing styles. This paper presents \system{}, an innovative text editor capable of identifying factual inaccuracies and suggesting corrections in over 90 languages. \system{} could potentially assist humans writers in content creation in sectors like news and media by helping editors detect factual errors early. However, end-to-end multilingual fact-checking presents unresolved challenges for both academia and the industry~\cite{Panchendrarajan:2024:arXiv}.

To make this problem tangible, our approach is threefold: First, we address the problem of detecting check-worthy claims, a task that involves understanding the context, relevance, and potential impact of each statement. Second, we address the task of generating and executing search engine queries, for gathering relevant information from the web. Finally, this information is then utilized by a Natural Language Inference (NLI) model, for veracity prediction. Furthermore, we use LLMs, to generate justification summaries and also suggest precise textual amendments for error rectification.

We also present preliminary evaluation results which show that a smaller transformer model, fine-tuned using datasets in small number of languages, can outperform large language models (LLMs) such as GPT-3.5-Turbo and Mistral-7b for both claim detection and veracity prediction tasks. On the other hand, LLMs excel at generative tasks such as 
summarization and suggesting claim corrections.

\section{Related work}
\label{sec:related}
\begin{figure*}[ht!!]
    \includegraphics[width=\linewidth]{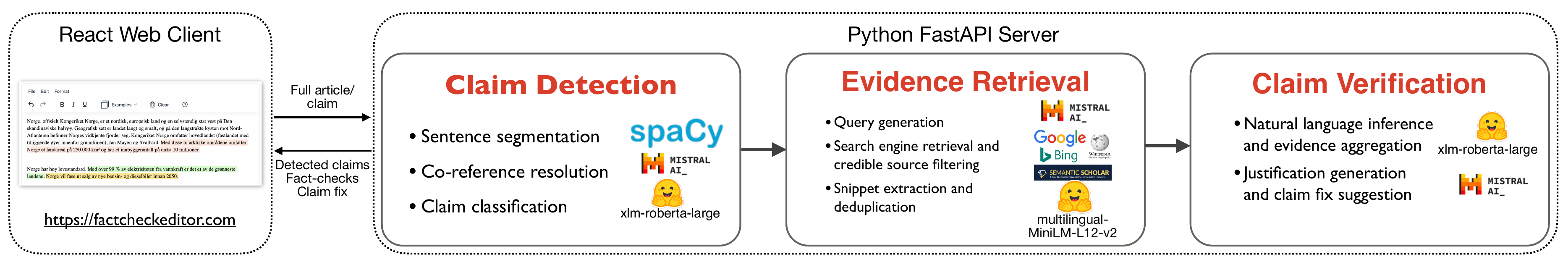}
    \caption{System Architecture of \system{}}
    \label{fig:system_arch}
\end{figure*}
Automated fact-checking has become popular in research recently. However, there is relatively low adaption in the industry. There are existing tools such as browser plugin proposed by~\citet{Botnevik:2020:SIGIRa} which can fact-check already written text. \citet{Wang:2023:arXiv} propose a tool for annotating the factual mistakes made by LLMs.  There are also tools for detecting hallucinations and factual mistakes made by LLMs, such as FactTool~\cite{Chern:2023:arXiva} and FAVA~\cite{Mishra:2024:arXiva}. While these are very sophisticated solutions, they do not focus on end-to-end fact-checking in a multilingual setting.

Majority of the fact-checking literature is focused on English language~\cite{Thorne:2018:arXiv,Guo:2022:JACL}. There are datasets for multilingual fact-checking \cite{Gupta:2021:ACL,Nielsen:2022:SIGIR}.
There is also recent survey on multilingual claim detection \citet{Panchendrarajan:2024:arXiv}. However, there is still a need for research regarding end-to-end multilingual fact-checking.

\section{System Overview}
\label{sec:method}

This work aims to provide a user-friendly web-based editor to compose textual articles with fact-checking feature. \system{}~identifies check-worthy claims in the written article by the user and verify those claims using evidence gathered from open web and previous fact-checks. Figure \ref{fig:system_arch}, the architecture of \system{}, with a web-based front-end implemented in React framework and a backend server. The frontend includes a text editor implemented using the TinyMCE text editor\footnote{\url{https://www.tiny.cloud}}. The backend, exposes REST APIs to interact with the machine learning (ML) models. The ML models used in the backend are grouped into (a) Check-worthy claim detection, (b) Evidence retrieval, and (c) Veracity prediction.

\subsection{Check-worthy Claim Detection}
\label{sec:claim_detection}
The goal of this stage is to identify and enrich claims which warrant verification.
\paragraph{Sentence segmentation and Co-reference resolution:}
The initial step in processing an article is sentence segmentation, which involves breaking down the text into individual sentences. We primarily use models from Spacy\footnote{\url{https://spacy.io/usage/models/}} for this task due to their efficiency and accuracy. We use an LLM (Mistral-7b), for co-reference resolutions, which helps in identifying the pronouns and linking them back to the appropriate named entities they represent.

\paragraph{Claim classification:}
This step is often the first step in a manual fact-checking pipeline. The objective is to determine whether a sentence contains a claim that warrants verification. We approach this as a binary classification task, where the goal is to classify each claim as either `check-worthy' or not. To accomplish this, we leverage established datasets ClaimBuster~\cite{Hassan:2017:VLDB} and CLEF CheckThat! Lab~\cite{Alam:2021:arXiv}.  These datasets are in English, therefore, we translate them to a handful of languages (Norwegian, German, and Danish) to fine-tune a multilingual classifier (\textbf{XLM-Roberta-Large}). Surprisingly, this limited training is able to transfer the knowledge to other languages, which the model didn't have any training data for. We also employ LLMs with \textbf{two-shot chain-of-thought (CoT) reasoning} prompts for comparison (See Section \ref{sec:claim_eval}).

\subsection{Evidence Retrieval}
    The goal of this stage is to retrieve highly relevant documents to verify the claims in the previous step. It is also important to retrieve both supporting and refuting documents for the claim. 

\paragraph{Query generation:}
This process involves generating effective questions or search queries to find relevant documents. Typically, original claims are used as search queries. However, this approach often fails to retrieve relevant documents, particularly for claims containing incorrect information. To address this, we draw inspiration from existing works~\cite{Fan:2020:arXivb,Schlichtkrull:2023:arXiva,Chen:2022:arXiva}. We utilize \textbf{Mistra-7b}\footnote{\url{https://ollama.ai/library/mistral}} to create more effective questions and queries that are well-suited for search engines. The prompts used can be found on our GitHub repo\footnote{\url{https://github.com/factiverse/factcheck-editor/tree/main/code/prompts}}.

\paragraph{Retrieval from search engines:}
We search across diverse platforms for queries and questions, including Wikipedia, Google Fact-check Explorer for previous fact-checks, Google and Bing search engines, and the semantic web for scholarly articles, ensuring a broad and comprehensive source coverage. To maintain source credibility, we filter out domains which are known to spread misinformation\footnote{\url{https://en.wikipedia.org/wiki/List_of_fake_news_websites}} and only consider scholarly articles from Semantic Scholar with at least 10 citations. To support multilingual search, we specify the language option in the respective search APIs.

\paragraph{Deduplication and Snippet Extraction}
To streamline search results from various sources, we implement deduplication by combining URL, title, and content, using approximate matching to filter out duplicates. Additionally, we refine relevance by extracting the top three paragraphs most related to the claim, determined by cosine similarity scores from sentence embeddings, focusing on the most pertinent information. We use \textbf{Multilingual-MiniLM-L12-v2} for computing sentence embeddings.

\begin{figure*}[h!!]
    \includegraphics[width=0.95\linewidth]{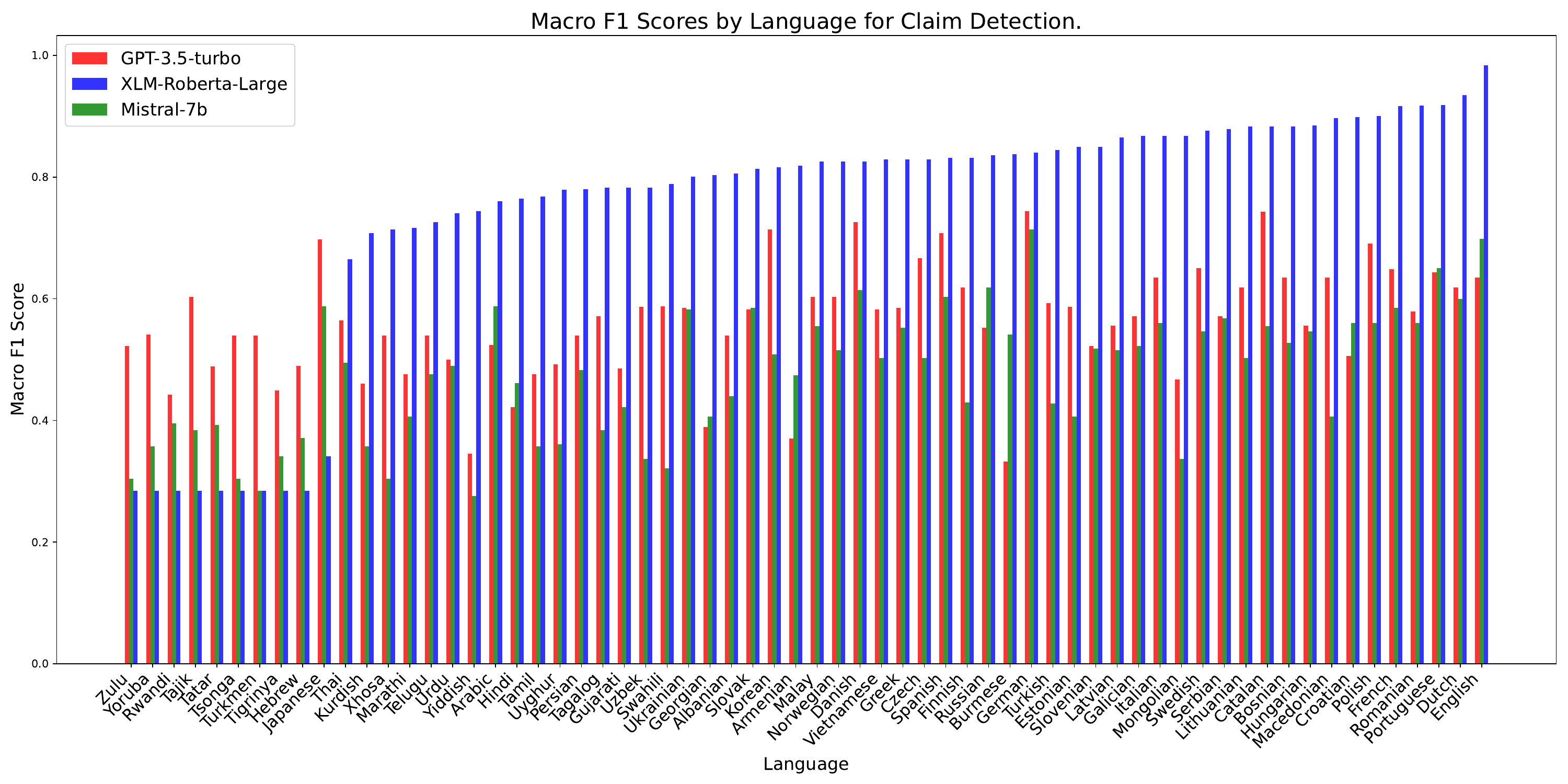}
    \caption{Evaluation of claim detection for 118 langauges usig XLM-RoBERTa-Large, GPT-3.5-Turbo and Mistral-7b.}
    \label{fig:claim_detection_macro}
\end{figure*}
\begin{figure}[h!!!]
    \includegraphics[width=0.9\linewidth]{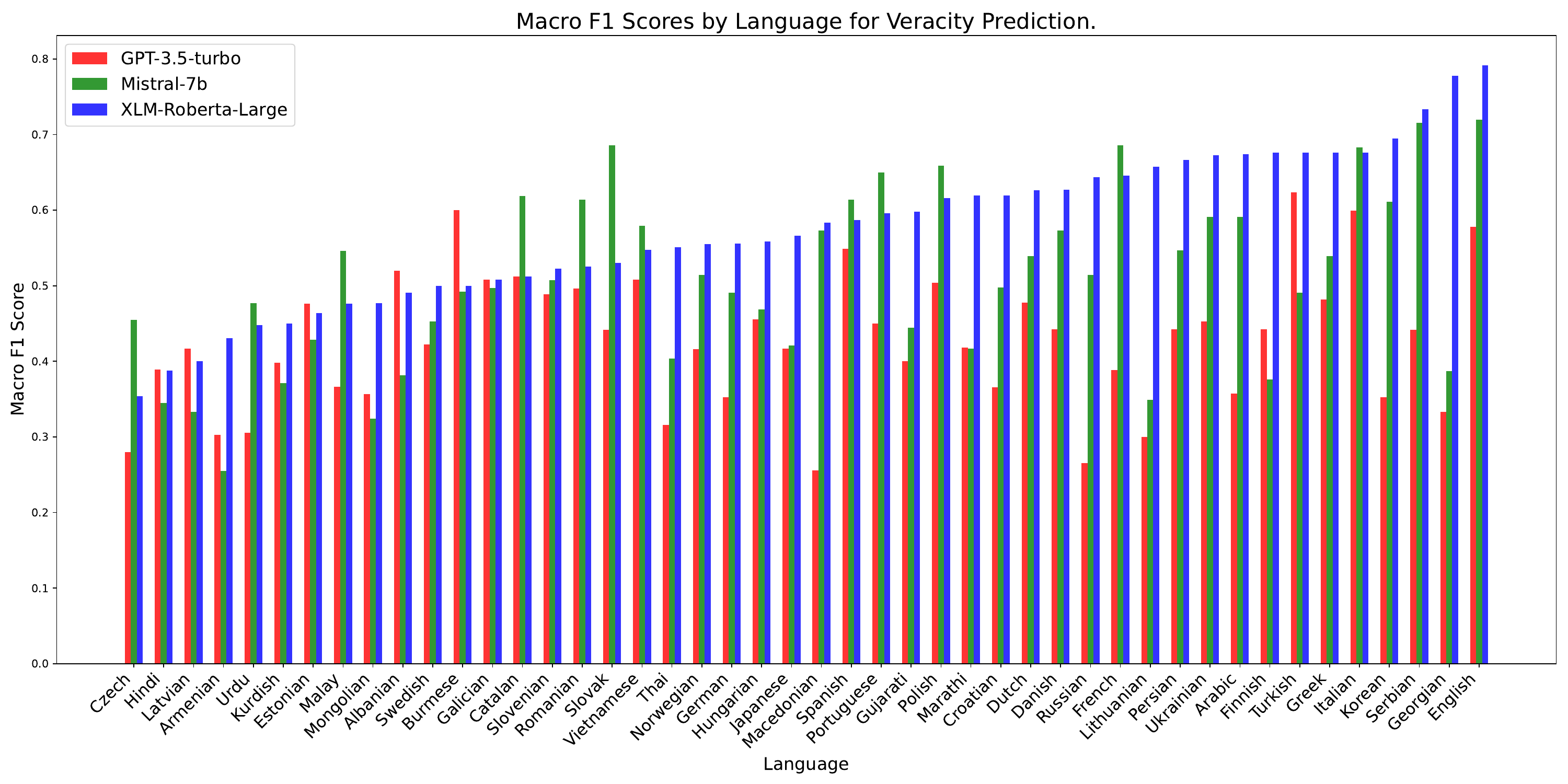}
    \caption{Evaluation of veracity prediction for 46 languages.}
    \label{fig:nli_eval}
\end{figure}
\subsection{Veracity Prediction}
    After having retrieved the evidence, and pre-processing them, the final step in the fact-checking pipeline is veracity prediction and justification generation. This is a crucial step in fact-checking.
\label{sec:veracity}
\paragraph{Natural Language Inference (NLI):}
The NLI task involves classifying whether a piece of evidence supports, refutes or unrelated to a claim~\cite{Bowman:2015:arXiv}. In our process, we ensure that only relevant documents are considered as evidence, significantly reducing the number of unrelated documents. 
Therefore, we simplify it by modeling it as a binary classification (supports or refutes). 
We fine-tune XLM-Roberta-Large using FEVER~\cite{Thorne:2018:arXiv}, MNLI~\cite{Williams:2018:arXiv} and X-Fact~\cite{Gupta:2021:ACL} datasets. We also compare its performance against LLMs from OpenAI and Mistral in Section \ref{sec:vercity_prediction}. 
For each claim, there are usually multiple evidence snippets, and the NLI prediction applies to each claim-evidence pair. To synthesize these individual predictions, we use the majority voting technique used in the literature \cite{Popat:2017:WWWa,Schlichtkrull:2023:arXiva}.

\paragraph{Justification Generation and Claim Fix Suggestion:}
With several evidence snippets available for a claim, each presenting different arguments that support or refute it, the information can become overwhelming for users. To enhance accessibility, we summarize the evidence in relation to the claim and its predicted veracity label, offering a concise and coherent overview that simplifies understanding the basis for the claim's classification. Similarly, in the way tools like Grammarly suggest corrections for typos and grammar, we introduce a method for suggesting fixes to inaccuracies in claims based on the evidence found. Utilizing Mistral-7b and crafted prompts, this feature not only identifies potential errors in claims but also offers suggestions for correction, thereby improving the accuracy and reliability of the information. We omit the qualitative evaluation of these suggestions as future work.

\section{Experimental Evaluation}
\label{sec:eval}

\subsection{Claim Detection}
\label{sec:claim_eval}
\paragraph{Dataset:} Fact-checking full articles is different from most existing datasets such as political debates, therefore, we annotate a smaller scale news dataset. The dataset statistics are described in Table \ref{tab:claim_detection}. Since this dataset is in English, we translated this dataset into 118 languages using the Google translate API.

\begin{table}
    \caption{Dataset distribution.}
    \centering
    \begin{tabular}{l|rrrrr}
    \hline
      \textbf{Split}   & \textbf{Not Check-} & \textbf{Check-} & \textbf{True} & \textbf{False} &\textbf{Total} \\
         & \textbf{worthy} & \textbf{worthy }& \textbf{Claims} & \textbf{Claims} & \\
      \midrule
       Train  & 609 & 548 & 332 & 196 & 1,076  \\
       Dev  & 38 & 25 & 15 & 10 & 63  \\
       Test  & 62 & 38 & 26 & 12 & 100   \\
       \hline
    \end{tabular}
    \label{tab:claim_detection}
\end{table}

\begin{table}
    \caption{Claim detection and veracity prediction results presented as mean Micro and Macro-F1 scores for all languages.}
    \centering
    \begin{tabular}{p{2.8cm}|rr|rr}
    \hline
      \textbf{Model}   & \multicolumn{2}{c}{\textbf{Claim Detection}}  & \multicolumn{2}{c}{\textbf{NLI}}  \\
         & \textbf{Ma.-F1} & \textbf{Mi.-F1 }& \textbf{Ma.-F1} & \textbf{Mi.-F1 } \\
      \midrule
       GPT-3.5-Turbo  & 0.562 & 0.567 &  0.427 & 0.461  \\
       Mistral-7b  & 0.477 & 0.510 & 0.509 & 0.557  \\
       XLM-RoBERTa-Large  & \textbf{0.743} & \textbf{0.768} & \textbf{0.575} & \textbf{0.594}   \\
       \hline
    \end{tabular}
    \label{tab:result_summary}
\end{table}
\begin{figure*}[h!!]
    \centering
    \includegraphics[scale=0.28]{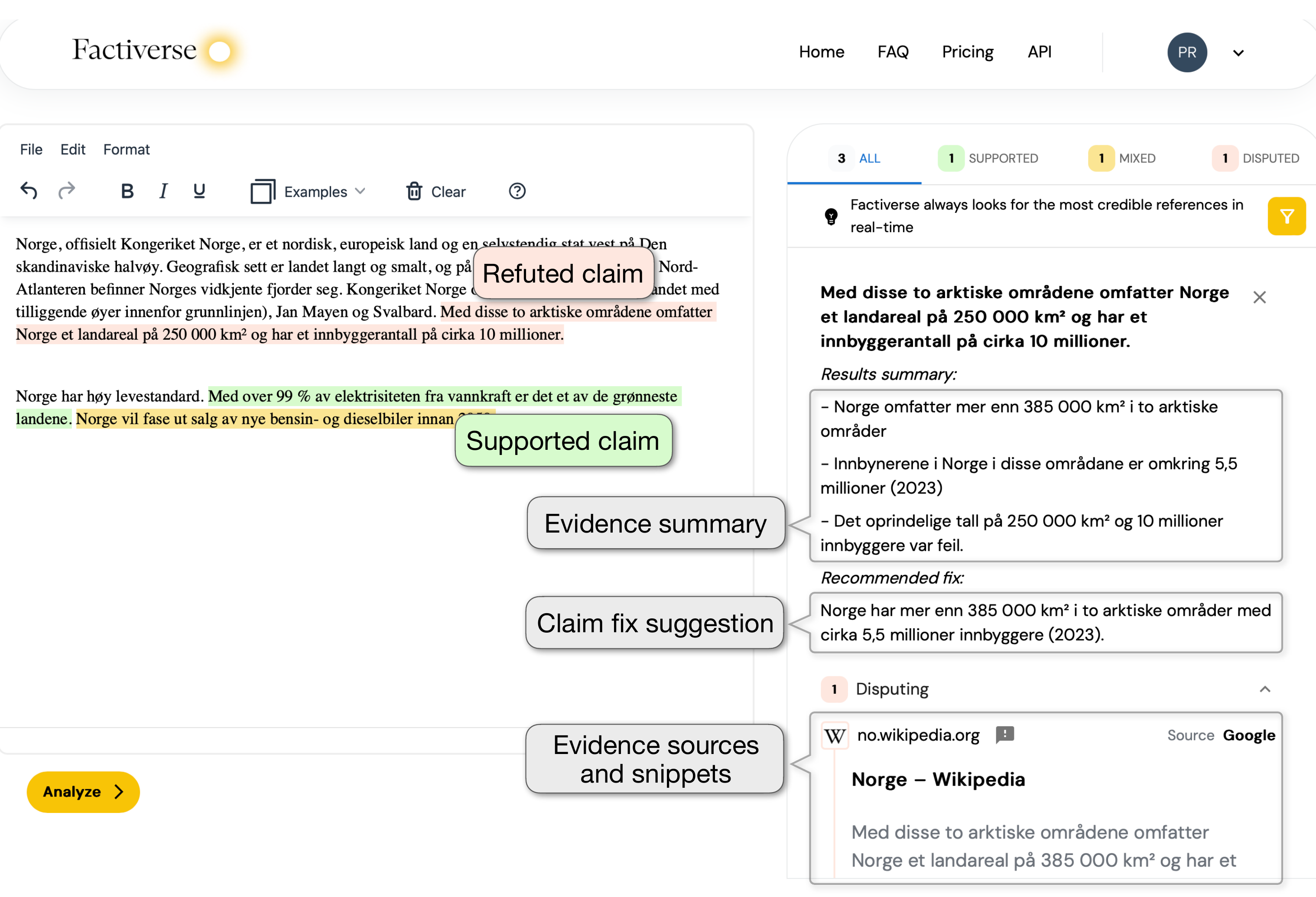}
    \caption{\system{}~demonstration}
    \label{fig:demo}
    \vspace{-0.25cm}
\end{figure*}

\paragraph{Results:} We compare XLM-RoBERTa-Large, GPT-3.5-Turbo and Mistra-7b using this dataset. In Figure \ref{fig:claim_detection_macro}, the F1-Macro is shown for all languages. Surprisingly, the fine-tuned XLM-RoBERTa-Large outperforms both GPT-3.5-Turbo and Mistra-7b in most languages. Since the model was trained mainly on English, not surprisingly it is the best performing language. For some languages, we see that XLM-RoBERTa-Large is the worst performing model. On closer inspection, these are the languages not supported (not included in the pre-training step). Mistral-7b seems to be the worst performing model, it seems to be because Mistral, struggles to follow instructions in the prompt. We observed a similar pattern in Micro-F1 scores, therefore, we omit those results due to lack of space. This suggests that for claim detection, fine-tuning a multilingual transformer model is promising rather than, few-shot chain-of-thought reasoning prompts with LLMs in a multilingual setting. Table \ref{tab:result_summary} shows the mean Macro-F1 and Micro-F1 scores for all languages evaluated. \textit{This shows that a fine-tuned XLM-RoBERTa-Large can outperform LLMs in multilingual setting for claim detection.}

\subsection{Veracity Prediction}
\label{sec:vercity_prediction}
\paragraph{Dataset:} We use the same data from the claim detection dataset for the NLI and veracity prediction tasks. The distribution of True and False claims are shown in Table \ref{tab:claim_detection}.

\paragraph{Results:} 

As shown in Figure \ref{fig:nli_eval}, fine-tuned XLM-RoBERTa-Large outperforms GPT-3.5-Turbo and Mistral-7b for most languages. It is interesting to see that Mistral performs better than GPT-3.5-Turbo despite being much a smaller LLM. Mistral-7b seems to be the best model for some European languages, such as French and Portuguese. Since for some languages, we couldn't find any evidence snippets for any of the claims, they are omitted. The overall results are shown in Table \ref{tab:result_summary} with similar observations to claim detection.

\section{Implementation and Demonstration}
\label{sec:concl}

We use a docker container to deploy the backend on a public cloud provider. Frontend is also hosted on the same provider. We use Ollama framework for self-hosting the LLMs. 

Figure \ref{fig:demo} shows an example involving an article written in Norwegian that contains factual inaccuracies. For instance, the claim ``Norge et landareal på 250 000 km$^2$ og har et innbyggerantall på cirka 10 millioner'' (which translates to ``Norway has a land area of 250,000 km$^2$ and a population of approximately 10 million'') is flagged as incorrect, with a suggestion to replace ``250 000'' with ``385 000'' and `10 million'' with ``5.5 million'' based on the evidence found. The editor also marks claims in red and `green to indicate disputed and supported claims, respectively, based on the evidence. Additionally, the right-hand pane displays evidence snippets along with a summary of the generated justification. The demonstration can be accessed live\footnote{\url{https://factcheckeditor.com}} and the evaluation code is shared\footnote{\url{https://github.com/factiverse/factcheck-editor}}.

\section{Conclusion}
In this paper, we demonstrated a multilingual text editor designed for identifying factual errors in written text. We also conduct preliminary experiments, which show that fine-tuning transformer models are more effective for claim detection and veracity prediction in multilingual setting with over 90 languages, warranting further research on end-to-end multilingual fact-checking.

\section{Acknowledgements}
This work is in part funded by the Research Council of Norway project EXPLAIN (grant number 337133). We also acknowledge Tobias Tykvart and Domante Stirbyt\`{e} from Factiverse for the implementation of the \system{} front-end.

\newpage
\bibliographystyle{ACM-Reference-Format}
\bibliography{sigir2024-ai-editor.bib}

\end{document}